\documentclass[]{bytedance_seed}

\usepackage{minitoc}
\usepackage{url}
\usepackage{amssymb}
\usepackage[utf8]{inputenc}
\usepackage{microtype}
\usepackage{booktabs}
\usepackage{pifont} 
\usepackage{multirow}
\usepackage{makecell}
\usepackage{paralist}
\usepackage{xspace}
\usepackage{color}
\usepackage{xcolor}
\usepackage{colortbl}
\usepackage{adjustbox}
\usepackage{hyperref} 
\usepackage[edges]{forest}
\usepackage{tikz} 
\usepackage{caption}
\usepackage{amsfonts}
\usepackage[toc,page,header]{appendix}
\usepackage[utf8]{inputenc}
\usepackage{bbm}
\usepackage{enumitem}
\usepackage{caption}
\usepackage{subcaption}
\usepackage{siunitx}
\usepackage{tabularray}
\definecolor{seedc}{RGB}{7, 92, 173}

\newcommand{\name}[1]{}
\newcommand{\hardware}[1]{}

\renewcommand{\paragraph}[1]{\vspace{0.1em}\noindent\textbf{#1}}

\title{Closing the Reality Gap: Zero-Shot Sim-to-Real Deployment for Dexterous Force-Based \\Grasping and Manipulation}

\author[]{ByteDance Seed}

\contribution[]{
Full Author List in Contributions}

\abstract{
Human-like dexterous hands with multiple fingers offer human-level manipulation capabilities, but training control policies that can directly deploy on real hardware remains difficult due to contact-rich physics and imperfect actuation. We close this gap with a practical sim-to-real reinforcement learning (RL) framework that utilizes dense tactile feedback combined with joint torque sensing to explicitly regulate physical interactions. To enable effective sim-to-real transfer, we introduce (i) a computationally fast tactile simulation that computes distances between dense virtual tactile units and the object via parallel forward kinematics, providing high-rate, high-resolution touch signals needed by RL; (ii) a current-to-torque calibration that eliminates the need for torque sensors on dexterous hands by mapping motor current to joint torque; and (iii) actuator dynamics modeling to bridge the actuation gaps with randomization of non-ideal effects such as backlash, torque–speed saturation. 

Using an asymmetric actor–critic PPO pipeline trained entirely in simulation, our policies deploy directly to a five-finger hand. The resulting policies demonstrated two essential skills: (1) command-based, controllable grasp force tracking, and (2) reorientation of objects in the hand, both of which were robustly executed without fine-tuning on the robot. By combining tactile and torque in the observation space with effective sensing/actuation modeling, our system provides a practical solution to achieve reliable dexterous manipulation. To our knowledge, this is the first demonstration of controllable grasping on a multi-finger dexterous hand trained entirely in simulation and transferred zero-shot on real hardware.
}

\date{22 December, 2025}

\checkdata[Project Page]{\url{https://dexmanip-seed.github.io/dexmanip}}

\begin{document}
\maketitle

\section{Introduction}
\label{sec:intro}

Reinforcement learning with sim-to-real has achieved remarkable success in locomotion community, leading to widespread industrial use cases of quadruped and humanoid robots. Pioneering works such as ANYmal’s agile locomotion~\cite{hwangbo2019learning} and soccer-playing humanoids~\cite{google2024soccer, tirumala2024learningrobotsocceregocentric} have demonstrated that a systematic design of domain randomization techniques, combined with RL techniques such as privileged learning of actuator dynamics and injection of noise in sensory observations, can bridge the reality gap for complex whole-body locomotion skills. Research in adaptive legged locomotion \cite{yang2020multi} further shows that while standard Multilayer Perceptron (MLP) architectures sufficed for learning individual narrow control policies, achieving complex multi-skill behaviors often necessitates the orchestration of a mixture of specialized neural network experts.

Regarding sim-to-real techniques, through methods like domain randomization, system identification, and learning actuator compensation model~\cite{hwangbo2019learning, lee2020learning, uan2025, he2025asap}, these technologies reduce the gap between the robot model itself and its interaction with the environment, enabling robots to move across complex terrains~\cite{highspeed, humanoidtransformer, hoeller2023anymal, cheng2023parkour, zhuang2023robot}, perform mobile manipulation~\cite{liu2024vbc, portela2023learningforce, ha2024umilegs, he2024legmanip, uan2025}, and mimic human motion~\cite{liao2025beyondmimicmotiontrackingversatile, he2025asap}. Recent advancements like ASAP~\cite{he2025asap} leverage residual action modeling to align simulated joint torques with real-world actuator responses, while UAN~\cite{uan2025} introduces unsupervised actuator networks to compensate for unmodeled nonlinear properties exhibiting in the real robots. 

These approaches share a common paradigm: (1) \textit{physics-based simulation augmentation} to cover hardware uncertainties (e.g., friction coefficients, motor saturation), (2) \textit{learning-based actuator modeling} to replace traditional model-based analytical methods which are overly simplified and inaccurate, and (3) \textit{RL learning techniques (asymmetric actor-critic learning and/or curriculum learning)} to inform critic with privileged information, or gradually train policies to increasingly realistic task conditions. Such techniques have enabled zero-shot deployment of dynamic skills like locomotion on uneven terrain and highly dynamic acrobatic flips, demonstrating the successful use of RL approach in deploying legged locomotion trained in simulation to reality in a zero-shot manner.

While sim-to-real RL has revolutionized legged locomotion, its application to \textit{dexterous in-hand manipulation} remains challenging and not fully addressed. Due to limitations in tactile sensors and motor modeling, the sim-to-real technology has yet to be widely applied in the field of manipulation. This is rooted in the discrepancy arising from two unique differences and complexity in manipulation: (1) \textit{contact-rich physics} involving multi-point, multi-contact interactions and the wide range of material variations in object (e.g., softness, deformations), which are difficult to model accurately in current physics simulations that are primarily catered for rigid body dynamics; and (2) \textit{sensorimotor coupling} requiring seamless and effective fusion of tactile feedback, joint positions and torques, and high-dimensional visual perception. Nevertheless, prior works in \textit{Touch Dexterity}~\cite{yin2023rotating} and \textit{Robot Synesthesia}~\cite{yuan2023robotsynesthesia} have shown that tactile-aware policies can achieve robust in-hand reorientation using binary sensors and point-cloud representations. 

These results suggest the huge potential of sim-to-real for dexterous manipulation, which is currently under-rated. Hence, we are motivated to address the sim-to-real gaps in \textit{contact modeling} and \textit{actuator dynamics} in a systematically manner. Drawing inspiration from legged robotics, we developed a holistic \textit{sim-to-real recipe} combining \textit{scalable tactile simulation}, \textit{current-to-torque calibration}, and \textit{actuator modelling with uncertainty in actuator dynamics} can close the perception-action loop for dexterous manipulation. This work represents an endeavor towards realizing human-level manipulation capabilities through physics simulation driven reinforcement learning.

In the grand vision of building general-purpose robots \cite{mccarthy2025towards}, multi-finger dexterous hands with anthropomorphic design have a unique advantage and great potential to offer human-like manipulation capabilities. However, their high degrees of freedom (DoFs) and complex, contact-rich dynamics present substantial control challenges. Consequently, most real-world robotic applications still rely on simple parallel-jaw grippers, and achieving robust \textit{in-hand manipulation}---the ability to reposition and reorient objects within a fixed grasp---remains a seminal challenge in robotics. The core difficulty lies in the need to sense and control intricate multi-point contact interactions, a domain where humans excel through the seamless integration of touch and proprioception.

Deep reinforcement learning (DRL) trained in simulation has emerged as a promising pathway to overcome the inherent complexities of contact-rich manipulation. The sim-to-real paradigm, leveraging large-scale domain randomization, has produced landmark results, such as transferring in-hand rotation policies to a real five-finger hand~\cite{andrychowicz2018learning}. This foundation was later extended to solve a Rubik's Cube~\cite{akkaya2019rubik}, establishing a robust methodology for bridging the reality gap through high-dimensional state spaces~\cite{tobin2017dr}. Recent efforts have further expanded the generalizability of this paradigm . Work such as~\cite{handa2023dextreme} demonstrated robust cube reorientation on an Allegro Hand through optimizing simulation fidelity and employing extensive randomization. 

A pivotal trend in this progress is the transition from pure vision or proprioception toward multimodal feedback. The integration of \textit{tactile sensing} has significantly enhanced manipulative precision and dexterity, enabling in-hand rotation using only dense binary touch signals~\cite{yin2023rotating}, fusing visuotactile and point-cloud inputs~\cite{yuan2023robotsynesthesia}, and facilitating the manipulation of thin slender objects~\cite{hu2025dexterous}. Complementing these advances, work on learning pre-grasping \cite{sun2020learning} shows that effective sim-to-real transfer is achievable for specialized single-purpose policies, demonstrating that functional dexterous behaviors can be acquired without reliance on manual heuristic programming.

Parallel progress in dexterous grasping~\cite{zhang2025robustdexgrasp} and contact-rich humanoid skills~\cite{lin2025humanoid} highlights that with appropriate sensing and modeling, sim-to-real RL is a powerful tool for learning policies deployable on real hardware. Furthermore, while a standard MLP is often sufficient for narrow control tasks, complex behaviors can be better facilitated by orchestrating multiple specialized experts. Recent hierarchical approaches \cite{roman2023hybrid} have further demonstrated that long-horizon manipulation skills can be realized by orchestrating a repertoire of learned skills. This suggests a compositional pathway for integrating individual skills into complex, long-sequential manipulation tasks. 

\begin{figure}[t]
    \centering
    \includegraphics[width=0.8\linewidth]{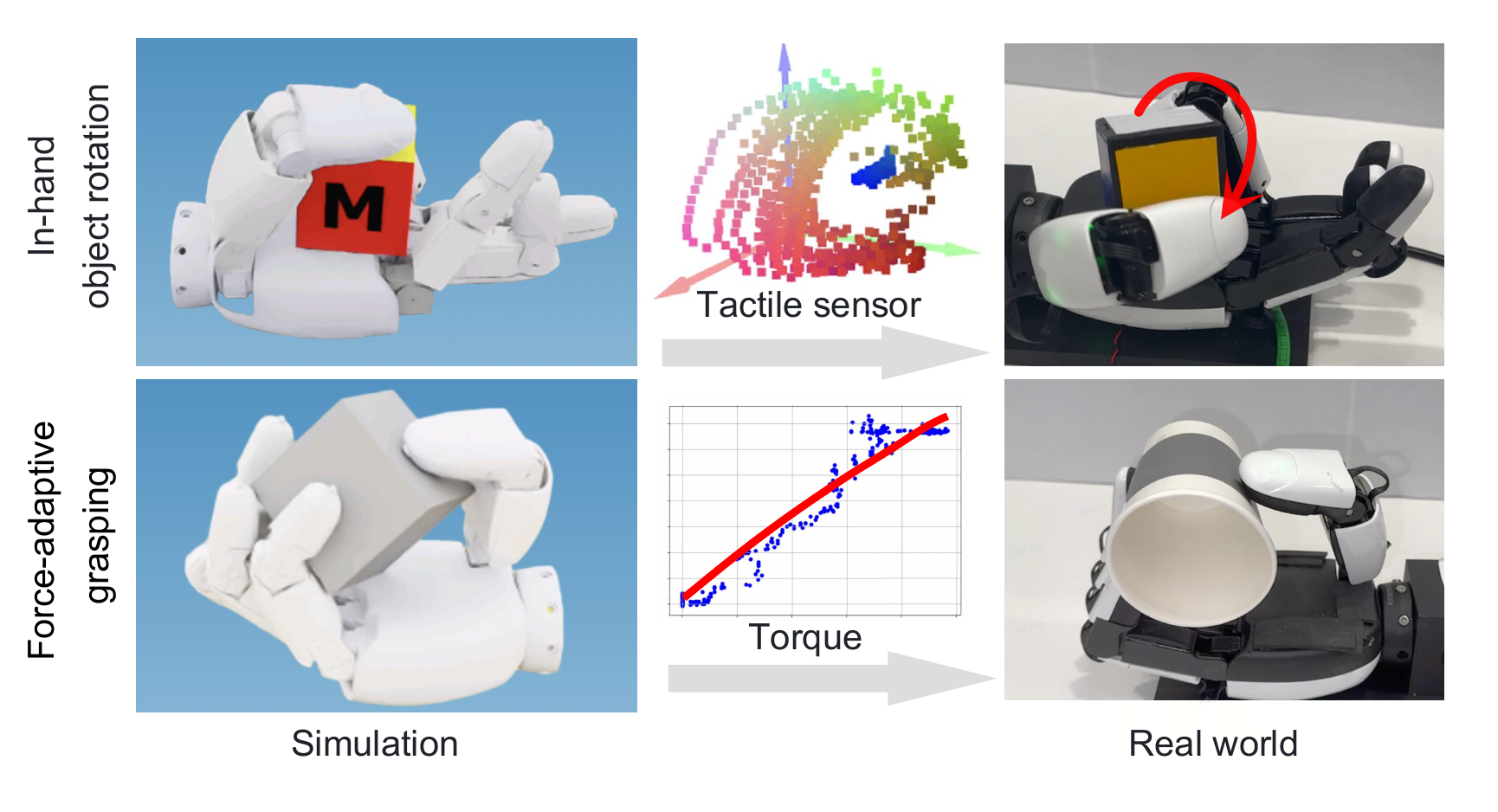}
    \caption{A framework for learning a full-state policy integrating tactile sensing and joint torques for dexterous grasping and in-hand manipulation.}
    \label{fig:framework}
\end{figure}

Despite these advances, a critical frontier remains largely unexplored: \textit{force-sensitive, closed-loop manipulation}. While prior work has excelled at \textit{position-centric} tasks (e.g., achieving a target orientation), the explicit regulation of interaction forces which is essential for handling fragile items or performing precision assembly, is often overlooked. Closing this gap necessitates policies that leverage \textbf{full-state feedback}, fusing high-resolution tactile sensing with joint-level torque control to perceive and command forces directly. As visualized in Fig.~\ref{fig:framework}, this ``touch \& torque'' observation space is the key to achieving reinforcement learning (RL) objectives for two fundamental skills: (i) \textbf{force-controllable grasping} for tracking commanded grip forces, and (ii) \textbf{in-hand reorientation} via controlled contact and slip.

However, the development of such full-state policies faces two profound practical hurdles that have prevented their widespread adoption:
\begin{enumerate}[topsep=2pt,itemsep=2pt,partopsep=2pt,parsep=2pt]
    \item \textbf{The Tactile Simulation Bottleneck:} Simulating high-resolution tactile contacts with high physical fidelity is computationally prohibitive at the scale required for deep RL. This forces a trade-off between simulation speed and accuracy, often resulting in brittle policies that have not sufficiently explored the contact dynamics encountered in reality.
    \item \textbf{The Actuation Reality Gap:} Most commercially available dexterous hands use semi-direct-drive actuation and lack joint-level torque sensors, relying instead on motor current for torque estimation. The reality gap between idealized actuator models in simulation (e.g., perfect torque control) and the non-ideal dynamics of real motors (e.g., torque-speed curves, friction, backlash) presents a major barrier to successful sim-to-real transfer for force-based tasks.
\end{enumerate}
In essence, discrepancies in simulating \textit{tactile contacts} and modeling \textit{mechatronic actuator properties} have prevented full-state policies from reaching their potential.

Motivated by the growing availability of high-DoF hands, this paper presents a reliable sim-to-real \textit{recipe} for learning robust full-state policies. Our method, which trains entirely in simulation and deploys zero-shot to a real five-finger hand, is built on the following contributions designed to address the aforementioned gaps:
\begin{itemize}[topsep=2pt,itemsep=2pt,partopsep=2pt,parsep=2pt]
    \item \textbf{Full-State Policy Formulation:} We design policy observations that jointly incorporate dense tactile signals and estimated joint torques, providing the necessary feedback for explicitly inferring contact states and regulating interaction forces.
    \item \textbf{Computationally Efficient Tactile Simulation:} We introduce a fast, high-resolution tactile simulation method that approximates contacts by computing distances between a dense array of virtual tactile units and the object via parallel forward kinematics. This offers the high-rate signals needed for RL without sacrificing critical contact information.
    \item \textbf{Data-Driven Actuator Modeling and Randomization:} We bridge the actuation gap through a calibrated current-to-torque mapping that eliminates the need for physical torque sensors. We further model and randomize non-ideal motor dynamics (e.g., torque-velocity curves), dramatically reducing sim-to-real torque discrepancies.
    \item \textbf{Zero-Shot Sim-to-Real Deployment:} We demonstrate the efficacy of our integrated approach by successfully deploying policies for two contact-rich skills on real hardware: (1) \textbf{controllable grasping} with commanded force tracking and (2) \textbf{in-hand object reorientation}.
\end{itemize}

In summary, this work provides an easy-to-replicate handbook for training full-state RL policies on a representative 12-DoF dexterous hand. By integrating tactile and motor-current sensing with computationally efficient simulation and robust actuator modeling, we achieve zero-shot sim-to-real transfer of advanced, force-controllable manipulation skills. To the best of our knowledge, this represents the first demonstration of such \textit{controllable grasping} with grip-force tracking and robust in-hand manipulation on a multi-finger hand, trained entirely in simulation and deployed without any fine-tuning.

\section{Related Work}
\label{sec:related_work}
We review research work closely related to our focused techniques in our sim2real pipeline for dextrous hands: (i) simulation of tactile sensors, (ii) modeling of motor dynamics, and (iii) current-to-torque mapping and estimation without direct joint torque sensors.

\subsection{Tactile Simulation}
High-fidelity, fast tactile simulation is a long-standing obstacle for learning contact-rich skills. For visuotactile sensors, recent systems libraries have pushed both realism and throughput. Akinola \textit{et al.} builds a library within Isaac Gym that synthesizes visuotactile images and contact-force distributions, and couples them with a policy-learning toolkit aimed at sim2real transfer~\cite{akinola2025tacsl}. In parallel, Nguyen \textit{et al.} integrate a soft-body FEM simulator with an optical visuotactile rendering pipeline inside Isaac Sim to capture elastomer deformation for GelSight-like skins~\cite{nguyen2024tacex}. Zhang \textit{et al.} focus on modeling multi-mode tactile imprints induced by different surface coatings/patterns and report high realism while remaining efficient enough for learning loops~\cite{zhang2025tacflex}. Beyond single-hand settings, works begin to exploit visuo–tactile simulation for complex, often bimanual, fine assembly: e.g., general sim2real protocols for marker-based visuotactile sensors~\cite{chen2024gpvt}, bimanual visuotactile assembly via simulation fine-tuning~\cite{huang2025vtrefine}.

On the policy side, tactile-only or visuo–tactile in-hand reorientation has seen rapid progress. Yin \textit{et al.}learns touch-only in-hand rotation on low-cost binary sensors, emphasizing robustness to sensing imperfections~\cite{yin2023rotating}. Yuan \textit{et al.} propose a point-cloud tactile representation to fuse vision and touch for in-hand rotation~\cite{yuan2023robotsynesthesia}. Purely tactile in-hand manipulation with a torque-controlled hand was shown in~\cite{sievers2022tactile}, while works also demonstrate DRL-based tactile control for slender cylindrical objects~\cite{hu2025dexterous}. For broader dexterous/bimanual touch, Lin \textit{et al.} studies bimanual tactile manipulation with sim-to-real deep RL~\cite{lin2023bitouch}. Collectively, these efforts motivate fast, high-fidelity tactile simulation and policy training; our distance-field–based tactile simulation targets this efficiency–fidelity trade-off specifically for RL.

\subsection{Motor Modeling}

For dexterous hands and manipulation, recent systems explicitly incorporate motor/drive limits or task-informed system identification. Huang \textit{et al.} employ privileged learning, system identification, and reinforcement learning to transfer functional grasps across diverse hands~\cite{huang2024fungrasp}. Tactile in-hand works with torque-controlled hands to build policies on top of explicit joint torque models~\cite{sievers2022tactile}. In non-prehensile manipulation, early sim-to-real studies showed the importance of accurate actuation/friction models plus ensemble dynamics to combat identification error~\cite{lowrey2018nonprehensile}.  

In our setting, we explicitly model the motor’s \textit{torque–speed} envelope~\cite{shin2023actuatorconstrainedreinforcementlearninghighspeed}, then \textit{randomize} these parameters (stall torque, speed constants, friction/ripple surrogates) to cover manufacturing tolerances and temperature/load variation. This follows the spirit of recent agile-mobility works where actuator models are aligned via residual learning or unsupervised actuator nets~\cite{he2025asap,fey2025uan}, but adapts them to the brushed-DC, gear-driven fingers typical of dexterous hands.

\subsection{Current-to-Torque Alignment}
Most dexterous hands available nowadays typically have their SDKs to send over measurements of {motor current} but lack direct torque sensors at the joint level of the robot. When torque feedback (or torque-conditioned policy inputs) is desired, estimating a reliable \textit{current$\to$torque} mapping becomes critical. In industrial manipulator literature, motor-current–based wrench estimation has a long history: Kalman filtering and momentum-observer formulations are used to estimate Cartesian forces and torques from joint currents and states~\cite{wahrburg2017tase,han2022sensorless}. 
Gold \textit{et al.} consider torsional deflection plus motor current to estimate joint torques~\cite{gold2019ifac}. 
From the calibration and control perspective, current-based impedance control explicitly fits actuator current-torque ratios and friction for current-controlled robots--eschewing force-torque sensors, bypassing the need for force-torque sensors~\cite{dewolde2024currentimpedance}. System identification methods that identify the models and parameters of the drive gains and dynamics to ensure that the modeled dynamics is physically consistent with the real measurements~\cite{gautier2014drivegains}.

In this work, we adapt and apply these principles to the small electric motors that are commonly in dexterous hands for the design of direct-drive or semi-direct-drive of the joints. We fit a ``current$\to$torque map'' under quasi-static conditions to capture effective torque constants and biases, accounting for the a minor loss of torque exertion due to the \textit{wear-and-tear} of the gear train. Empirically, this mapping provides sufficiently accurate torque estimates for the RL policies and improves sim-to-real alignment,without physically building and embedding expensive miniaturized joint-torque sensors in the robot joints.

\section{Methodology}
\label{sec:method}

This section presents the technical details of our framework for learning two essential dexterous manipulation policies in simulation for zero-shot deployment on real hardware. We first formalize the problems in the RL framework for force-adaptive and controllable grasping, plus the in-hand object reorientation. Then, we describe our approach of designing the policy's full-state observation space, which includes both the tactile sensing and joint torque sensing via motor-current approximation. Finally, we delineate the integrated \textit{recipe} for the sim-to-real transfer, and explain technical knowhows on how we bridge the critical gaps in tactile sensing, contact physics, and actuator dynamics.

\subsection{Problem formulation}
This work addresses the \textbf{sim-to-real transfer} of dexterous manipulation policies for multi-fingered robotic hands, focusing on zero-shot deployment of simulation-trained policies onto real hardware. Despite progress in RL-based manipulation, transferring contact-rich, force-sensitive tasks, such as variable-force grasping and in-hand reorientation, remains challenging due to gaps in tactile simulation, actuator modeling, and contact physics.

Formally, given a dexterous hand with $N$ fingers equipped with tactile sensors and current-controlled motors (no torque sensors), we learn a policy $\pi:o_t\mapsto a_t$ that maps observations in simulation that perform successfully on the physical system without fine-tuning.

\subsection{MDP formulation}

We formulate the control of a dexterous hand as a Markov Decision Process (MDP), defined by the tuple $(S, A, T, R, \gamma, \rho_0)$, where $S$ is the state space, $A$ is the action space, $T(s_{t+1} | s_t, a_t)$ is the state transition probability, $R(s_t, a_t)$ is the reward function, $\gamma$ is the discount factor, and $\rho_0$ is the initial state distribution. Our objective is to learn a policy $\pi_{\theta}(a_t | o_t)$, parameterized by $\theta$, that maximizes the expected cumulative reward $\mathbb{E}{\pi}[\sum_{t=0}^{T} = \gamma^t R(s_t, a_t)]$.

The core challenge we address is the \textbf{sim-to-real transfer} of this policy. While the policy is trained entirely within a simulated MDP $M_{\text{sim}}$, where $M_{\text{sim}}$ represents the \textit{simulated physics world} -- the entire simulation environment including robot model, objects, and sensor / contact models. The training data is generated by the policy $\pi$ interacting with $M_{\text{sim}}$, and the challenge is that such a policy trained in $M_{\text{sim}}$ must execute successfully or at least perform reasonably well on the \textit{real physical world}, governed by a real-world MDP $M_{\text{real}}$, under a zero-shot deployment --- meaning no fine-tuning on the real hardware. 

The discrepancy between $M_{\text{sim}}$ and $M_{\text{real}}$, known as the ``reality gap'', is particularly pronounced for dexterous manipulation due to:
\begin{itemize}
    \item Tactile Simulation Gap: Inaccurate or computationally prohibitive modeling of high-resolution contact sensing;

    \item Actuator Dynamics Gap: Simplified models of motor dynamics that omit effects like torque-speed saturation, backlash, and current-to-torque nonlinearities;

    \item Contact Physics Gap: Differences in friction, material deformation, and multi-body contact dynamics between simulation and reality.
\end{itemize}

Formally, our agent is a dexterous hand with $N$ fingers ($N=5$ in this paper with 12-DoF active actuation). Its actions $a_t \in A \subset \mathbb{R}^{12}$ are target joint positions for the 12-DoF hand, sent to a motor-level PD controller. The observations $o_t \in O$ constitute a full-state representation, including proprioception (joint angles, velocities), tactile readings, and estimated joint torques (detailed in Section~\ref{subsec:observation}).

The policy $\pi_{\theta}(a_t | o_t)$ is trained in simulation to solve two distinct, contact-rich tasks:
\begin{itemize}
    \item Force-Adaptive Grasping: The policy needs to achieve a stable grasp and modulate the \textit{each finger's grip force} to track a user-specified force command $F_{\text{cmd}}$.

    \item In-Hand Object Reorientation: The policy needs to rotate an in-hand object about a fixed axis through coordinated finger movements, keeping stable contact throughout the whole multi-contact interactions.
\end{itemize}

The success of our approach relies not only on the policy's architecture, but also on the careful design of $M_{\text{sim}}$ to minimize the reality gap. The subsequent sections detail our methods for bridging the gaps in sensing (Section~\ref{subsec:tactile_sim}) and actuation (Section~\ref{subsec:actuation_model}).

\subsubsection{Force-Adaptive Grasping}
The objective of this task is to achieve and maintain stable grasps on a diverse set of objects with geometrically complex and physically varied properties including size, shape, mass distribution, and surface friction, while accurately exerting user-specified force levels. The policy must modulate grip forces in real time according to object characteristics and external disturbances to prevent slip or excessive deformation, thereby demonstrating robust and adaptive grasping under dynamic real-world conditions.

\subsubsection{In-Hand Object Rotation}
This task requires the precise and controlled rotation of a grasped object about a predefined spatial axis using continuous and compliant finger gaits. Successful rotation entails rich, multi-point frictional contacts and requires coordinated finger motions that induce controlled slip and rolling interactions between fingertips and the object surface. The policy must maintain stability during manipulation by adjusting contact forces and reposing fingers as needed to avoid dropping or losing control of the object, thereby achieving smooth and continuous rotational reorientation.

Experiments use a 12-DoF direct-drive dexterous hand (xHand)~\cite{robotera2024generic} and are conducted in the IsaacLab simulation environment~\cite{mittal2023orbit}.


\subsection{Identification of Sim-to-Real Gaps}
Our study identified 4 primary sim-to-real gaps:

\textit{(1) \textbf{Perceptual Gap}:} Simulations often provide perfect ground-truth state information, such as object pose and joint angles. However, the real systems rely on these states from noisy sources, including errors from camera calibration, varying lighting conditions, occlusions, and the limitations of vision-based pose estimation, resulting in an unavoidable mismatch between the policy's ideal expected observations and the actual ones.

\textit{(2) \textbf{Discrepancies in Actuator Dynamics}:} Real actuators, such as the commonly used electric motors, have complex and non-ideal behaviors that are often simplified or omitted in the physics simulation. It ranges from significant mechanical discrepancies in the physical gear chain system, such as mechanical backlash in gear transmissions, to nonlinearities such as static (stiction) and dynamic friction, torque-speed curves (e.g., motor current saturation at higher velocities), motor response delays, and imprecise current-to-torque mapping. These unmodeled acutator dynamics cause the commanded actions in reality exert different forces and thus exhibits different motion patterns on the real robot.

\textit{(3) \textbf{Discrepancies in Contact Physics}:} The contact dynamics in the simulation is primarily based on rigid bodies, which often have limited accuracy in surface geometry due to the use of simplified meshes, and also miss out the material properties (e.g., stiffness, friction, restitution) which are more complex to numerically compute. Therefore, for ensuring reasonable computational cost,  to some extent, most available simulation engines nowadays trade off the complexity and fidelity of real physical interactions, including multi-contact modeling, deformation of soft and compliant materials, varying friction coefficients, rolling resistance, and the stiffness-damping properties of the building materials of the robot. All these result in policies that fail to effectively explore and exploit rich tactile feedback in simulation, which is essential for dexterous in-hand manipulation.

\textit{(4) \textbf{Unseen State Distribution Shift}:} Given the above main sources of discrepancies, even with robust training with randomization of key physical properties, a policy may still encounter novel object properties such as materials, shapes, masses, frictions--scenarios and configurations that are not exactly the same or not seen during training. This Out-of-Distribution (OOD) problem leads to compounding errors as the policy operates in those states where its learned value functions and policies are no longer accurate, causing performance deterioration or even failures while deployed in the real system.

\subsection{Training setup for Reinforcement Learning}

\subsubsection{Force-adaptive grasping} 

A critical factor in training effective RL policies is the quality and efficiency of data collection within the simulation environment. For dexterous manipulation, the chosen training scenario profoundly impacts both the speed of policy convergence and the ultimate quality and robustness of the grasp.

Several noticeable \textit{limitations} of traditional tabletop grasping are summerized as follows:

\textbf{Conventional RL training for dexterous hands often mimics human tabletop grasping}: an object is placed on a surface, and the hand, initialized from a random pose above it, must learn to approach, orient, and grasp the object. While intuitive, this paradigm introduces significant inefficiencies and challenges:

\textbf{High Sample Inefficiency and Complex Reward Engineering}: 
A randomly initialized policy must discover this complex sequence of actions (approaching, palm reorientation, and finger closure) through extensive trial-and-error. This exploration process is exceptionally time and computationally expensive. It often necessitates intricate curriculum learning and meticulously engineered reward functions to incrementally guide the policy toward successful behavior, adding considerable complexity to the training pipeline.

\textbf{Reward Hacking and Non-Robust Emergent Behaviors}: To maximize the reward, policies frequently learn to exploit the simulation-specific physics that are inconsistent with the real-world physics, resulting in policies that fail to translate to reality with reasonable success rate -- a phenomenon known as reward hacking (typically, the unrealistic frictions). This can result in unnatural, unstable, and aesthetically non-human-like grasping strategies (e.g., precarious two-finger pinches or exploiting specific contact points). These strategies are often brittle and fail catastrophically upon real-world deployment due to the inevitable sim-to-real gaps in dynamics and contact physics, rendering the trained policy ineffective.

To overcome these limitations, we introduce a novel training paradigm: an inverted ``catch-the-object'' setup. As illustrated in Figure~[\ref{fig:grasp}], the dexterous hand is fixed in a palm-up orientation. Objects are procedurally dropped from above into its workspace, and the policy's objective is to robustly catch and stabilize them.

This method is implemented with several key randomization strategies to enhance robustness and generalization.

\textbf{Object Properties}: Mass, size, shape, and friction coefficients are randomized.

\textbf{Initial Conditions}: Introduced randomization to create variations of the initial orientation of different objects, before releasing them into the palm of the dexterous hand.

\textbf{Inverted Catching Setup-Learning Grasping via Catching}: The \textit{Inverted Catching} setup turns the hand upside-down and drops different objects into the hand with an open-palm configuration, using gravity to naturally pull the object towards the hand with varied velocities and poses, making the RL policy easier to learn how to grab and hold the targeted object firmly. In such a setting, the robot avoids the pre-grasp interactions which can lead to hacking the physics simulation.

The proposed approach offers several distinct advantages that directly address the shortcomings of traditional methods:\\
\textbf{Improved Sample Efficiency}: By leveraging gravity to naturally bring the object into the hand's workspace, the exploration problem is vastly simplified. The policy can focus its learning effort on the core challenge of finger coordination and force modulation upon contact, rather than on the initial approach phase. This leads to a significant reduction in training time and computational resources.

\textbf{Mitigation of Reward Hacking and Emergence of Human-Like Grasps}: The catching dynamic encourages the formation of enveloping, multi-point contact grasps that naturally center the object in the palm. This inherently suppresses the emergence of unnatural, reward-hacking strategies and promotes stable, human-preferred grasping configurations. The resulting policies are not only more robust but also more aesthetically plausible.

\textbf{Enhanced Sim-to-Real Transferability}: The observations required for this task are proprioception (joint angles, velocities), inferred torque, and tactile contact signals, which can be primarily measured on real hardware. By avoiding reliance on hard-to-simulate or hard-to-measure visual features (like precise object pose from a specific camera angle) and fostering robust contact-rich interaction, this paradigm minimizes the perceptual and dynamics sim-to-real gaps. Consequently, simulation-trained policies can be deployed very successfully on physical robots in a zero-shot manner.

In summary, this inverted catching setup serves as a highly efficient and effective ``gymnasium'' for training policies for grasping various objects, which accelerates learning and encourages robust grasping strategies, and fundamentally reduces the reality gap, serving as a practical handbook for achieving reliable sim-to-real dexterous manipulation.

\subsubsection{In-Hand Object Rotation}
The in-hand object rotation task is designed to test the dexterous hand's ability to perform fine, contact-rich manipulation. A standardized cube is chosen as the manipulation object to provide a structured and reproducible benchmark. This allows us to focus on the core challenge of learning coordinated finger gaits rather than adapting to infinite object geometries.

Constrained Rotation for Focused Learning:
Given the limited degrees of freedom (DoF) of the adopted dexterous hand compared to the human hand, we constrain the agent to rotate the cube around a single, predefined axis. This simplification reduces the problem's dimensionality, allowing the policy to master the fundamental dynamics of controlled slipping and rolling contacts without being overwhelmed by the complexity of full 3D rotation. Mastery of single-axis rotation is a critical prerequisite for more general manipulation.

To encourage the policy to learn a continuous, sustainable manipulation skill rather than just a single, static reorientation, we implement a dynamic training curriculum. The task is structured as a series of sequential goals:
The target pose for the policy is not a fixed absolute orientation. Instead, it is defined as a 90-degree rotation from the object's current orientation around the designated axis.

Each time the policy successfully achieves a 90° rotation (within a specified tolerance), this event is recorded as a success. The environment then automatically updates the target to a new one, another 90° further along the same axis.

This ``shifting goal'' mechanism forces the policy to learn a continuous regrasping behavior. It must not only achieve a specific orientation but also recover stability and prepare for the next manipulation step, mimicking the continuous nature of in-hand manipulation performed by humans. This is far more challenging and informative than learning a one-shot rotation from a fixed start to a fixed end pose.

\subsection{Policy learning in simulation}
\label{subsec:observation}

\begin{table}[]
    \begin{subtable}[t]{0.495\linewidth}
    \centering
    \begin{tabular}{@{}c|ccc@{}}
    \toprule
    Input                  & Dim & Actor                     & Critic                    \\ \midrule
    Hand joints angles     & 12D            & \checkmark & \checkmark \\
    Hand joints torque     & 12D            & \checkmark & \checkmark \\
    Object position        & 3D             & \checkmark & \checkmark \\
    Object linear velocity & 3D             & \checkmark & \checkmark \\
    Contact force          & 5D             & \checkmark & \checkmark \\
    Contact center         & 15D            & \checkmark & \checkmark \\
    Fingertip positions    & 15D            & \checkmark & \checkmark \\
    Force command          & 1D             & \checkmark & \checkmark \\ \bottomrule
    \end{tabular}
    \caption{Observation for force-adaptive grasping}
    \label{tab:RewardGrasping}
    \end{subtable}
    \begin{subtable}[t]{0.495\linewidth}
        \centering
        \begin{tabular}{@{}c|ccc@{}}
        \toprule
        Input                       & Dim & Actor & Critic \\ \midrule
        Hand joints angles          & 12D            &  \checkmark     &  \checkmark     \\
        Relative target orientation & 6D             &  \checkmark     &  \checkmark      \\
        Last actions                & 12D            &  \checkmark     &  \checkmark      \\
        Contact center              & 15D            &  \checkmark     &  \checkmark      \\
        Contact force               & 5D             &  \checkmark     &  \checkmark      \\
        Fingertip positions         & 15D            &  \checkmark     &  \checkmark      \\
        Object orientation          & 6D             & ×     &  \checkmark      \\
        Fingertip velocities        & 30D            & ×     &  \checkmark      \\
        Fingertip rotations         & 30D            & ×     &  \checkmark      \\
        Hand joints velocities      & 12D            & ×     &  \checkmark      \\
        Object linear velocity      & 3D             & ×     &  \checkmark      \\
        Object angular velocity     & 3D             & ×     &  \checkmark      \\ \bottomrule
        \end{tabular}
        \caption{Observation for in-hand object rotation}

        \label{tab:obs_rotate}
    \end{subtable}
    \caption{Observation for our tasks}
\end{table}

\begin{table}[]

\end{table}

\subsubsection{Force-adaptive grasping} 
The complete set of observations used during policy training is detailed in Table~\ref{tab:RewardGrasping}. In the real-world deployment setup, the object’s three-dimensional position is estimated through a vision-based pipeline: the Z-coordinate is computed from the distance between the palm center and the centroid of the segmented object mask, where the mask is inferred using the Segment Anything Model (SAM)~\cite{kirillov2023segment}. The X and Y coordinates are held fixed relative to the hand’s initial pose, simplifying the state estimation under the assumption of limited lateral motion during grasping. The object’s linear velocity is then numerically derived from the sequence of estimated Z-coordinates using a finite difference method. Additionally, a force command scalar in the range [0, 1] is provided as part of the observation, indicating the desired grasping intensity, enabling the policy to modulate grip force according to task requirements.
During the simulation training process, we designed a reward mechanism to guide the policy toward achieving stable and reliable grasping during inverted catching tasks.

Once the fingers are in contact with the object, we employ torque command rewards $R_{\text{torque}}$ to encourage the agent to apply specific joint torques. The target torque $\tau_{\text{target} = \tau_{\text{max}} \cdot F_{\text{cmd}}}$ is a continuous value provided by the instruction of the environment, and the reward function has a special handling for the thumb compared to the other fingers.

The total reward for this component is the sum of the individual finger torque rewards, which is applicable only when the finger is in contact with the object.
\begin{equation}
R_{\text{torque}} = w_{\text{torque}} \cdot \sum_{i \in \text{fingers}} R_{\text{torque}, i} \cdot I_{\text{contact}, i},
\end{equation}where $w_{\text{torque}}$ is the weight of this reward, $I_{\text{contact}, i}$ is the indicator function, which is 1 if the finger $i$ is in contact with the object and 0 otherwise.

The reward for the thumb is binary, contingent on its torque being within a valid operational range defined by the hardware limits $[\tau_{min}, \tau_{max}]$:
\begin{equation}
R_{\text{torque, thumb}}=I(\tau_{min}\le\tau_{thumb}\le\tau_{max}),
\end{equation}

Other four fingers' reward is a Gaussian function centered around the instructed target, multiplied by a validity mask.
\begin{equation}
R_{\text{torque}, i} = \exp\left(-\frac{(\tau_i - \tau_{\text{target}})^2}{2\sigma^2}\right) \cdot I(\tau_{min}\le\tau_{thumb}\le\tau_{max}) ,\quad
\end{equation}
where $\sigma$ is a tunable parameter determined by the desired torque tracking precision.

\begin{figure}

    \centering
    \includegraphics[width=0.9\linewidth]{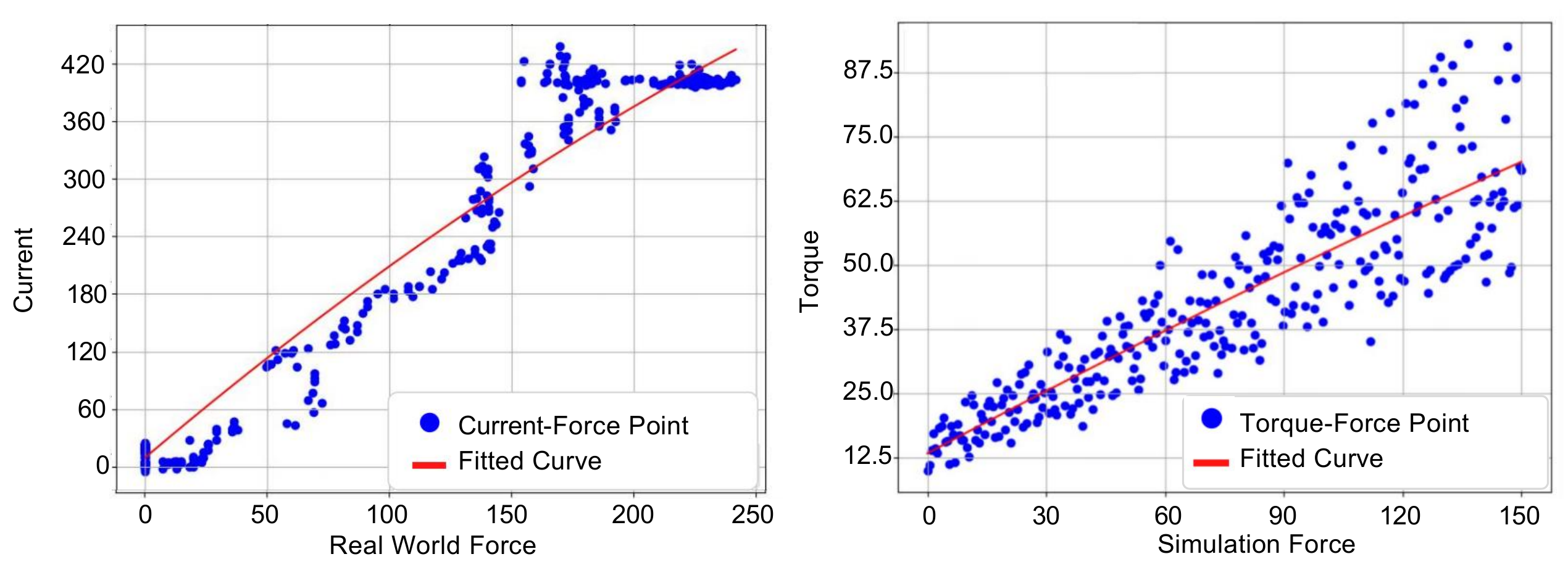}
    \caption{Calibration and alignment of \textit{current-force} (real robot) versus \textit{torque-force} (simulation) properties.}
    \label{fig:fig_force}

\end{figure}
For the contact force, similar to the joint torque reward, an appropriate contact force $F_{\text{target}} = F_{\text{cmd}} \cdot F_{\text{max}}$ is encouraged.
\begin{equation}
R_{\text{force}} = w_{\text{force}} \cdot \sum_{i \in \text{fingers}} R_{\text{force}, i} \cdot I_{\text{contact}, i},
\end{equation}
\begin{equation}
    R_{\text{force, thumb}} = I(F_{min} \le F_{\text{thumb}} \le F_{max}),
\end{equation}

\begin{equation}
R_{\text{force}, i} = \exp\left(-\frac{(F_i - F_{\text{target}})^2}{2\sigma^2}\right) \cdot I(F_{min} \le F_i \le F_{max}) ,\quad
\end{equation}
where $F_i$ is the fingertip contact force of the root joint (finger $i$), $\sigma$ is the standard deviation of the Gaussian function, and $I(\cdot)$ is the indicator function.

In addition to standard grasping rewards, we introduce a novel four-finger consistency reward to promote coordinated motion among the four homologous and structurally similar fingers. This term regulates the uniformity of their flexion and extension angles. In the human hand, biomechanical constraints limit the range of independent motion of the metacarpophalangeal joints, causing adjacent fingers to move correlatively. In contrast, robotic fingers are driven by independent motors. Without explicit coordination, reinforcement learning policies often produce unnatural and inefficient postures. The consistency reward encourages more human-like motion patterns, leading to more stable and physically plausible grasps with improved force distribution.

The penalty $R_{\text{diff}}$ to quantify the difference in-between four fingers is defined as:
\begin{equation}
  R_{\text{diff}} = w_{\text{diff}} \cdot \text{Var}(\{q_j \mid j \in \text{inner fingers}\})  ,
\end{equation}
where $w_{\text{diff}}$ is the penalty weight, $q_j$ is position of the first joint of finger j, $\text{Var}(\cdot)$ is variance of the set of joint positions.

The outer finger movement penalty $R_{\text{outter}}$ is defined as:
\begin{equation}
    R_{\text{outter}} = w_{\text{outter}} \cdot \|{q_{\text{outter}}} - {c_{\text{outter}}}\|_2,
\end{equation}
where $w_{\text{outter}}$ is the weight of this penalty, ${q_{\text{outter}}}$ is current joint position vector for the outer joints (index, middle, ring, and little fingers), ${c_{\text{outter}}}$ is center position vector of the outer joints.

In addition, a set of standard penalty terms is integrated to further shape the agent’s behavior. To discourage early termination of episodes, a terminal state penalty is introduced. Moreover, an action rate penalty is applied to limit large fluctuations in action commands between consecutive time steps, thereby promoting smoother and more stable control policies. This penalty is formally defined as:
\begin{equation}
R_{\text{action}} = w_{\text{action}} \cdot \|{a}_t - {a}_{t-1}\|_2^2,
\end{equation}
where $R_{\text{action}}$ is the action rate penalty, $w_{\text{action}}$ is the weight of this penalty, ${a}_t$ is the action vector at the current time step $t$, and ${a}_{t-1}$ is the action vector at the previous time step $t-1$.
And a joint velocity L2 penalty to discourage high joint velocities. This encourages the agent to generate smoother and more stable movements, avoiding abrupt motions. The penalty is defined as:
\begin{equation}
R_{\text{vel}} = w_{\text{vel}} \cdot \|\dot{{q}}\|_2^2,
\end{equation}
where $\dot{{q}}$ represents the joint velocity vector, and $w_{\text{vel}}$ is the corresponding penalty weight.

\subsubsection{In-Hand Object Rotation}

\begin{table}[]

\centering
\begin{tabular}{@{}c|c|c@{}}
\toprule
Reward & Formula & Weight \\ \midrule
Close to goal & $ \frac{1.0}{|d_{\text{rot}}| + \epsilon}\times \frac{1}{1 + e^{\alpha (d_{\text{goal}} - \delta_{\text{dist}})}}$ & $w_{\text{goal}}$ \\
Action penalty & $||{{a}_t - {a}_{t-1}}||^2$ & $w_{\text{action}}$ \\ \midrule
Reset reward & Condition & Value \\ \midrule
Reach goal bonus & $d_{\text{rot}}<\delta_{\text{rot}}^{\text{threshold}}$ and $d_{\text{pos}}< \delta_{\text{pos}}^{\text{threshold}}$ & $C_{\text{Bonus}}$ \\ \bottomrule
\end{tabular}
\caption{Reward for in-hand object rotation}

\label{tab:cube_reward}
\end{table}

Effective sim-to-real transfer requires policy observations that are not only informative but also practically obtainable on real hardware. We therefore design our policies around two feasible perception paradigms, as detailed in Tab.~\ref{tab:obs_rotate}:

\textbf{Proprioception-Only Policy:} This policy relies solely on the robot's internal state, joint angles, velocities, tactile signals ($F_i$, $\mu_i$), and actions, without any explicit object pose information. It must infer the object's state and rotation progress implicitly from the history of contact interactions, making it robust to the absence of external sensors but potentially more challenging to train.

\textbf{IMU-Augmented Policy:} This policy incorporates direct orientation feedback from a low-cost inertial measurement unit (IMU) embedded within the object. This provides a clear, real-world measurable signal of the object's state, simplifying the policy's task. The target orientation is specified relative to the robot's palm frame.

A critical implementation detail for the IMU-augmented policy is the choice of orientation representation. We avoid quaternions due to their double-cover property (where a single rotation is represented by both $\mathbf{q}$ and $-\mathbf{q}$), which introduces a discontinuity that can severely disrupt gradient-based learning. Instead, we adopt the continuous 6D rotation representation \cite{zhou2019continuity}, which provides a smoother learning landscape and leads to more stable and efficient policy convergence.

Reward Engineering for Continuous Manipulation: Simply rewarding the agent for minimizing the rotation error to a target can lead to a suboptimal local minimum: the policy learns to hold the object perfectly still to avoid any position penalty, thus never attempting the rotation. To overcome this, we designed a composite reward function (Tab.~\ref{tab:cube_reward}) that jointly minimizes both rotation error and position drift. Furthermore, a large bonus is provided upon successfully reaching a target orientation. This combination incentivizes the policy to perform complete rotations and then re-stabilize the object, emulating the continuous finger-gaiting motions seen in human dexterous manipulation. This reward structure was essential for promoting dynamic and continuous in-hand rotation rather than static grasping.

\subsection{Sim-to-Real Transfer}

\begin{figure}

    \centering
    \begin{subfigure}[b]{0.48\linewidth}
        \centering
        \includegraphics[width=\linewidth]{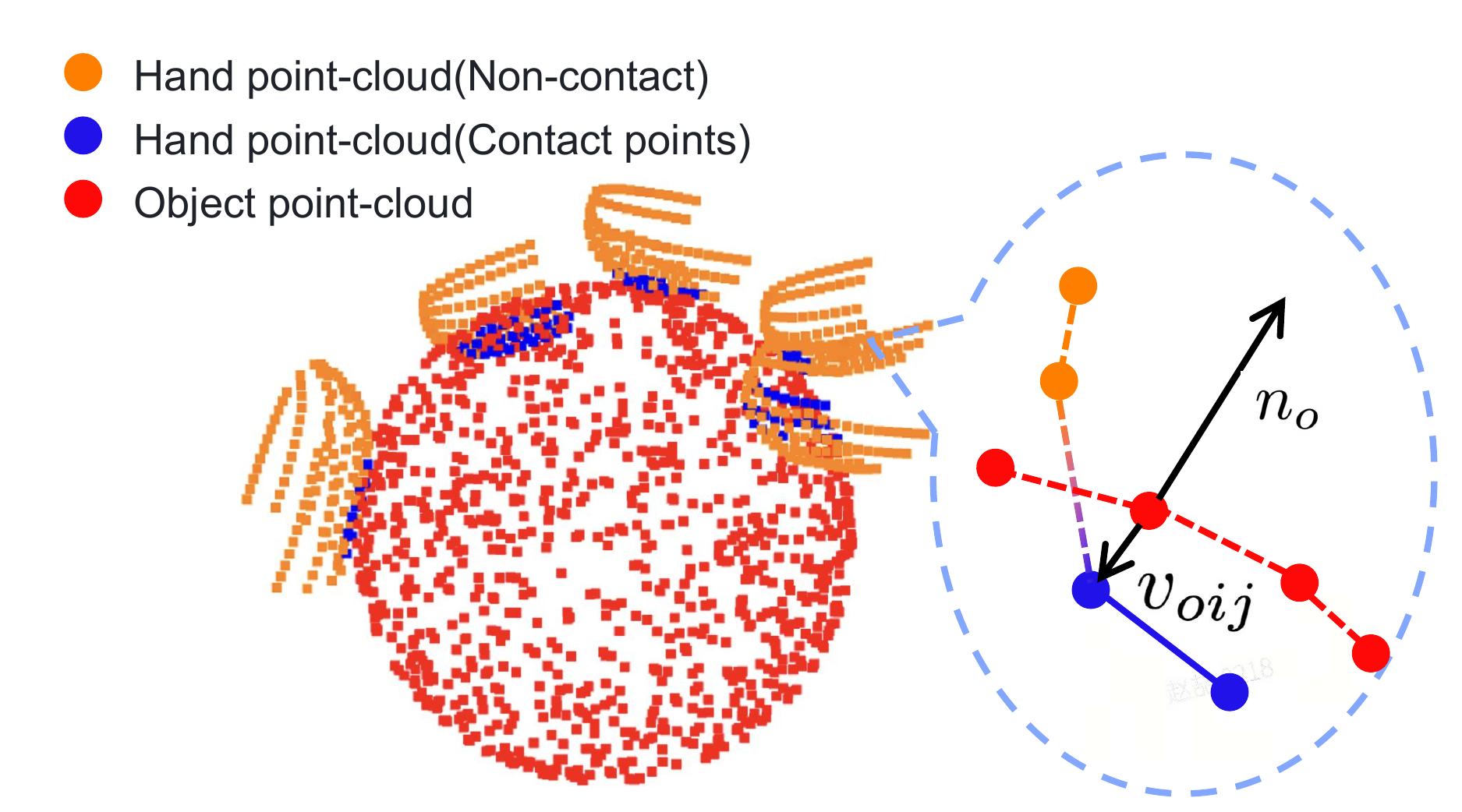}
        \caption{Modeling of contact points to approximate tactile sensors that mitigate the sim-to-real gap.}
        \label{fig:contact}
    \end{subfigure}
    \hfill
    \begin{subfigure}[b]{0.48\linewidth}
        \centering
        \includegraphics[width=0.9\linewidth]{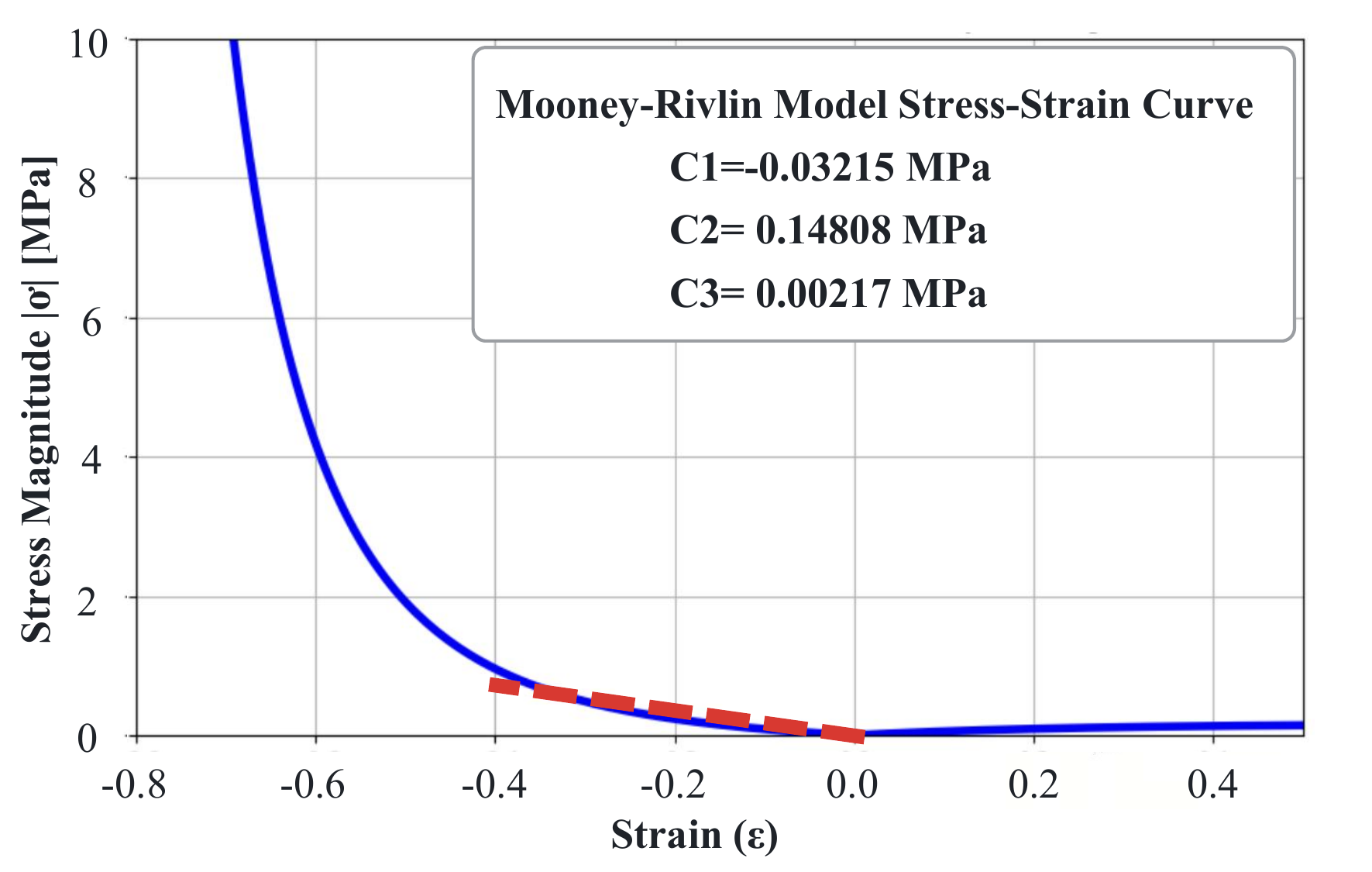}
        \caption{Stress-Strain curve of the rubber materials of the tactile sensor on the dexterous hands.} 
        \label{fig:Stress-Strain}
    \end{subfigure}
    \caption{Contact point modeling and material properties.}
    \label{fig:combined}

\end{figure}

We identified three primary contributors to the sim-to-real gap in our task: tactile sensing, complex in-hand contact simulation, and actuator dynamics. Our approach closes this gap through precise problem formulation and tailored simulation techniques.

\subsubsection{Tactile Sensor Simulation}

We implement a rapid, massively tactile sensor simulation using parallel forward kinematics, which calculates the distance from each sensor unit to the N-nearest surface points on the object.

We consider a dexterous hand with \(N\) fingers. Typically, \(N = 5\). The sensor set of the entire dexterous hand can be denoted as \(S\). On the fingertip of each finger \(i\) (where \(i \in \{1, 2, \ldots, N\}\)), there is a set of sensors, which we represent as \(S_i\). Therefore, the total sensor set is the union of all fingertip sensor sets:
\[ S = \bigcup_{i = 1}^{N} S_i. \]
For each individual sensor \(j\) on each fingertip \(i\) (where \(j \in \{1, 2, \ldots, M_i\}\), and \(M_i\) is the total number of sensors on the \(i\) - th fingertip), we define its key attributes:

\textbf{Position}: Through forward kinematics, we can calculate the spatial coordinates of each sensor from the hand degrees-of-freedom position, denoted as \(p_{ij} \in \mathbb{R}^3\).

\textbf{Activation}: On the real robot hand, the sensor is activated while reading is greater than a threshold. In the simulation, as shown in Fig.~\ref{fig:contact}, the nearest point ${p}_o$ on the surface of the object is found for each sensor ${p}_{ij}$. A vector from the object pointing to each \textit{tacxiel}, ${v}_{oij}$, is defined. The contact is detected if the sensor has deformed, which is determined by evaluating the dot product of ${v}_{oij}$ and the normal vector of the object at ${p}_o$. The vector from ${p}_o$ to ${p}_ij$ is given by:${v}_{oij} = {p}_{ij} - {p}_o$. Let ${n}_o$ be the outward-pointing normal vector at the object point ${p}_o$. If the dot product of ${v}_{oij}$ and ${n}_o$ is negative, it implies that the sensor has penetrated the object's surface, indicating a deformation. All points meeting such a condition are considered as \textit{equivalent contact points}.
    \begin{equation}
         \text{Contact} \iff {v}_{oij} \cdot {n}_o < 0 .
    \end{equation}

\textbf{Force}: On the real robot, normal forces are measured by the sensors. In the simulation, it is approximated using Mooney-Rivlin stress: strain relationship model to quantify the contact force of each sensor based on the distance $d_{ij} =|v_oij|$ between the sensor and the object surface and the total contact force per finger, denoted as \(f_{ij} \in \mathbb{R}^+\).

In our model, we use the fully actuated robotic hand xHand with 12 degrees of freedom. Each fingertip has \(M_i = 120\) sensors, so the entire hand has \(\sum_{i = 1}^{N}M_i = 600\) sensors. These sensors are concentrated in five independent ``dense groups''. This high local resolution on each fingertip can capture fine pressure distributions.

\subsubsection{Complex Contact Simulation}
\label{subsec:tactile_sim}

Simulating complex multi-point contacts in dexterous manipulation is a primary challenge. To bridge the sim-to-real gap for contact physics, we employ domain randomization on key object properties during policy training, including surface friction, damping, and restitution coefficients. This forces the policy to learn robust strategies that are invariant to these physical uncertainties, rather than overfitting to precise but inaccurate simulated dynamics.

To efficiently process the high-dimensional output of our simulated tactile sensors, we abstract the raw data into a compact, semantically meaningful representation $T_i$ for each fingertip $i$. This representation is designed to mirror the information a policy could feasibly extract from real tactile sensors, balancing detail with computational tractability. We define $T_i$ as a tuple capturing the two most critical aspects of contact: its magnitude and its location.

\begin{equation}
T_i = (F_i, {\mu}_i)
\end{equation}

\textbf{Total Contact Force ($F_i$):} This scalar value represents the magnitude of the interaction. On the real robot, this is the sum of normal forces from all individual taxels on the fingertip. In simulation, it is computed equivalently as the sum of the forces $f_{ij}$ from all active virtual sensors $j$ on the fingertip:
\begin{equation}
F_i = \sum_{j \in S_i^*} f_{ij}.
\end{equation}
We model the contact force $f_{ij}$ at each tactile unit using a general hyper-elastic material model to capture the non-linear relationship between the penetration distance $d_{ij}$ and reaction force. Here, the Mooney-Rivlin model is employed as the standard and generic formulation for such soft contacts. Specifically, because the rubber material used in the xHand fingertip is relatively rigid, under the common contact forces, the force-penetration curve for this material operates primarily in a low-nonlinearity and near-linear range relevant to typical manipulation forces, as shown in Fig.~\ref{fig:Stress-Strain}. Therefore, for computational efficiency during the training of reinforcement learning, we use a linear spring approximation of the contact force in the simulator: $f_{ij} \approx k \cdot d_{ij}$.

\textbf{Force-Weared Mean Contact Position (${\mu}_i$):} This 3D vector represents the center of pressure, a vital cue for estimating contact geometry and object pose. It is calculated as the force-weighted average of the positions ${p}{ij}$ of all active sensors:
\begin{equation}
{\mu}_i = \frac{\sum{j \in S_i^*} f{ij} \cdot {p}{ij}}{F_i}.
\end{equation}
Substituting $f_{ij} \approx k \cdot d_{ij}$ simplifies this calculation in simulation, making it highly efficient without loss of generality:
\begin{align}
      {\mu}_{i,sim} &\approx  \frac{\sum_{j \in S_i^*} d_{ij}\cdot k \cdot {p}_{ij}}{D_i\cdot k}\nonumber =\frac{\sum_{j \in S_i^*} d_{ij}\cdot {p}_{ij}}{D_i} .
\end{align}

\begin{figure}[t]
    \centering

    \includegraphics[width=1.0\linewidth]{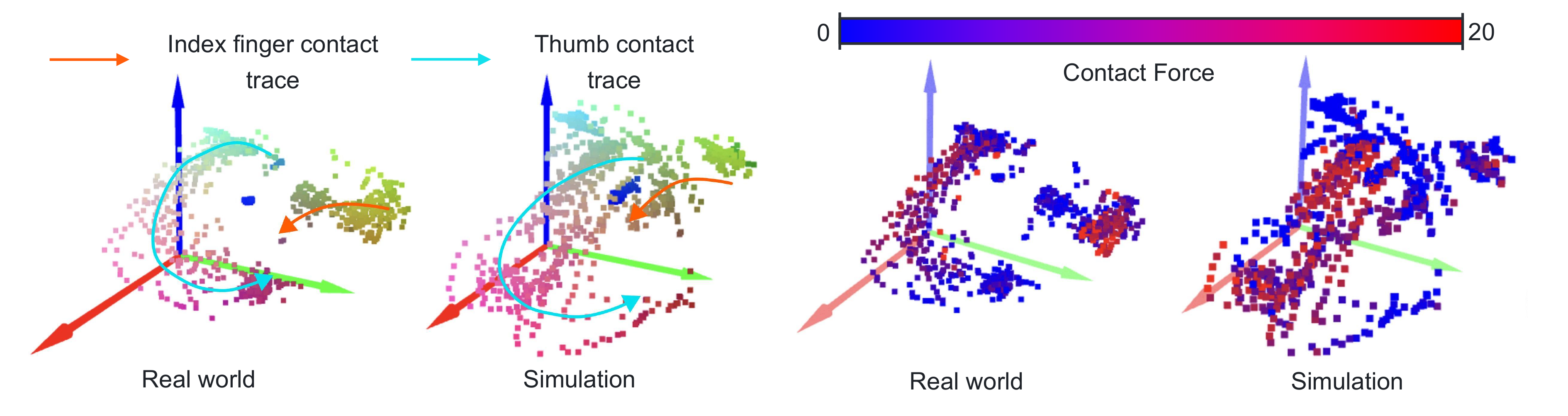}
    \caption{Visualization of real-world and simulated contact data during an in-hand rotation task. The close alignment between the contact points (Top) and contact forces (Bottom) shows the high fidelity of our contact simulation.}
    \label{fig:contact_point}
\end{figure}

This abstraction serves a dual purpose: it drastically reduces the observation space dimensionality for the RL policy, and it provides a transferable feature representation. As validated in Fig.~\ref{fig:contact_point}, the distributions of $F_i$ and ${\mu}_i$ are highly consistent between simulation and the real world. Crucially, the simulated contact region envelops the real-world data, confirming that the policy learns from a sufficient and covering distribution of contact scenarios, enabling effective sim-to-real transfer of tactile-based manipulation skills.

\subsubsection{Current-Torque Calibration}
A critical sim-to-real gap arises from the fact that our real-world dexterous hand provides motor current measurements but lacks direct joint torque sensors, whereas our simulation has direct access to idealized joint torques. To enable our torque-conditioned policy to function on real hardware, we must establish a reliable mapping from measured motor current to the simulated joint torque that the policy expects.

We bridge this perception gap through a one-time, per-joint calibration procedure. For each finger joint, we applied a spectrum of external forces to the fingertip in both the real and simulated environments. In the real world, we recorded the motor current $I_{\text{real}}$ and the resultant contact force $F_{\text{real}}$ measured from the high-resolution tactile sensors on the fingertip. In parallel, within simulation, we recorded the idealized joint torque $\tau_{\text{sim}}$ and the simulated contact force $F_{\text{sim}}$ for the same applied force conditions.

Our analysis, summarized in Fig.~\ref{fig:fig_force}, confirmed a strong linear correlation between contact force and both motor current (in the real world) and joint torque (in simulation). This linearity allows us to model the relationships as:

\begin{equation}
F_{\text{real}} \approx \alpha \cdot I_{\text{real}} \quad \text{and} \quad F_{\text{sim}} \approx \beta \cdot \tau_{\text{sim}}
\end{equation}

where $\alpha$ and $\beta$ are the estimated proportionality constants for the real and simulated systems, respectively.

We then identified the maximum observed values: the maximum real-world current $I_{\text{max}}$, the maximum simulated torque $\tau_{\text{max}}$, and their corresponding maximum contact forces $F_{\text{max, real}}$ and $F_{\text{max, sim}}$. The final calibration step involves normalizing both the real-world current and simulated torque signals into a unified, dimensionless force proxy scale of $[0, 1]$:

\begin{align}
  \quad & I_{\text{norm}} = \frac{I_{\text{real}}}{I_{\text{max}}} \approx \frac{F_{\text{real}}}{F_{\text{max, real}}} \\
 \quad & \tau_{\text{norm}} = \frac{\tau_{\text{sim}}}{\tau_{\text{max}}} \approx \frac{F_{\text{sim}}}{F_{\text{max, sim}}}
\end{align}

This normalization effectively aligns the real-world current reading with the simulated joint torque value through their shared relationship to contact force. During real-world deployment, the policy now receives $I_{\text{norm}}$ in the ``joint torque'' field of its observation space. This signal is semantically consistent with the $\tau_{\text{norm}}$ it was trained on in simulation, enabling zero-shot transfer without requiring physical joint torque sensors. This process ensures that a policy command for a specific ``simulated torque'' results in a functionally equivalent ``real-world force'' exertion.
\subsubsection{Actuator Model Randomization}
\label{subsec:actuation_model}

To bridge the sim-to-real gap in dexterous manipulation, we use a randomized actuator model that captures real-world motor nonlinearities, such as torque-velocity saturation and backlash hysteresis, improving policy robustness against hardware variations. The torque is computed via a PD controller with randomized gains:
\begin{equation}
\tau_c = k_p \cdot \left( q_{\text{ref}} - q_m \right) + k_d \cdot \left( \dot{q}_{\text{ref}} - \dot{q}_m \right),
\end{equation}
where $k_p$ and $k_d$ are the proportional and derivative gains, respectively, and $q_m$ denotes the measured joint position.

A key feature of the model is the incorporation of backlash, simulating mechanical dead zones due to gear play. The effective torque is modulated by a deadband function:
\begin{equation}
\tau_b = 
\begin{cases} 
0 & \text{if } |q_{\text{ref}} - q_m| < \epsilon \\
\tau_c & \text{otherwise}
\end{cases},
\end{equation}
where $\epsilon$ is the backlash threshold, randomized during training to mimic physical wear and tolerance variations.

Further, the torque is constrained by a velocity-dependent saturation function due to the DC motor's characteristics:
\begin{align}
\tau_{\text{sat}}^{+}(\dot{q}) = \tau_0 \left(1 - \frac{|\dot{q}|}{\dot{q}_{\text{max}}}\right), \hspace{2mm}
\tau_{\text{sat}}^{-}(\dot{q}) = -\tau_0 \left(1 - \frac{|\dot{q}|}{\dot{q}_{\text{max}}}\right),
\end{align}
where $\tau_0$ is the stall torque and $\dot{q}_{\text{max}}$ is the maximum no-load velocity. The final applied torque is given by:
\begin{equation}
\tau_{\text{applied}} = \eta \cdot \text{clip}\left( \tau_b,\ \tau_{\text{sat}}^{-}(\dot{q}),\ \tau_{\text{sat}}^{+}(\dot{q}) \right),
\end{equation}
where $\eta$ is a randomized factor that accounts for variations in motor torque constant and drive efficiency.

All model parameters are resampled for each actuator at the beginning of every training episode. This comprehensive randomization strategy forces the control policy to adapt to a wide spectrum of actuator imperfections, significantly improving sim-to-real transfer performance.

\section{Experimental Results}
\label{sec:exp}

\subsection{Force-adaptive grasping}

\begin{table}[t]

\centering
\begin{tabular}{@{}cc|cc@{}}
\toprule
$R_{\text{Force}}$      & $R_{\text{torque}}$     & Contact force range & Joint torque percent  \\ \midrule
\checkmark & \checkmark &  \numrange[range-phrase ={\,$\sim$\,}]{44.50}{93.92}                   &         \numrange[range-phrase ={\,$\sim$\,}]{1.09}{1.52}               \\
\checkmark & ×           & \numrange[range-phrase ={\,$\sim$\,}]{37.29}{70.02}                     &        \numrange[range-phrase ={\,$\sim$\,}]{0.69}{0.75}                    \\
          × & \checkmark &      \numrange[range-phrase ={\,$\sim$\,}]{38.45}{47.12}               &            \numrange[range-phrase ={\,$\sim$\,}]{0.85}{1.02}               \\ \bottomrule
\end{tabular}

\caption{Ablation on the rewards $R_{\text{Force}}$ and $R_{\text{torque}}$. }
\label{tab:abl_force}
\end{table}

\begin{figure}[t]
    \centering
    \includegraphics[width=1\linewidth]{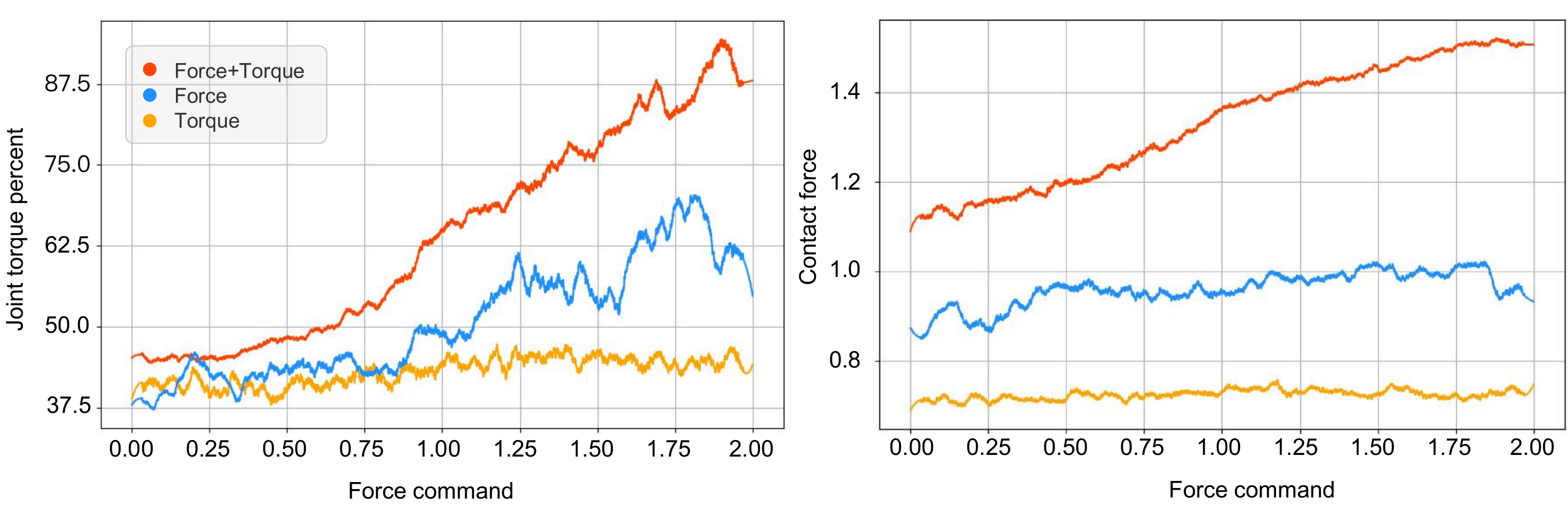}
    \caption{Joint torque and contact forces under controllable force commands with different reward settings.}
    \label{fig:Force_command}
\end{figure}

\begin{table}[t]
\centering
\caption{Ablation analysis of multiple observation combinations in the real world.}
\label{tab:abl_cube}
\resizebox{\linewidth}{!}{%
\begin{tabular}{|>{\centering}m{0.07\linewidth}|>{\centering}m{0.07\linewidth}|>{\centering}m{0.07\linewidth}|>{\centering}m{0.121\linewidth}|>{\centering}m{0.237\linewidth}|>{\centering\hspace{0pt}}m{0.079\linewidth}|>{\centering\hspace{0pt}}m{0.073\linewidth}|>{\centering}m{0.094\linewidth}|>{\centering\arraybackslash}m{0.104\linewidth}|} 
\hline
Contact center & Contact \par{} force & Force\par{} weighted & Orientation \par{} representation & Cons. success trials(sorted) & Average       & Median        & Succ.\par{} time(Ave.) & Time to fall (Ave.)  \\ 
\hline
                      \checkmark&\checkmark&\checkmark& 6D                                & 8,10,12,15,19,25,26,35,46,55 & \textbf{25.1} & \textbf{22.0} & 3.36                   & \textbf{84.3}              \\ 
\hline
                      \checkmark&\checkmark&\checkmark& Quaternion                        & 1,1,1,1,2,3,4,5,5,6          & 2.9           & 2.5           & 5.24                   & 15.2                       \\ 
\hline
                      \checkmark&\checkmark& ×                    & 6D                                & 3,6,8,10,11,13,13,15,21,22   & 12.2          & 12.0          & 3.75                   & 45.7                       \\ 
\hline
                      \checkmark& ×                    &\checkmark& 6D                                & 2,6,10,11,13,14,16,18,21,21  & 13.2          & 13.5          & 3.19                   & 42.1                       \\ 
\hline
                      \checkmark& ×                    & ×                    & 6D                                & 1,3,6,6,7,8,9,11,11,13       & 7.5           & 7.5           & 4.03                   & 30.2                       \\ 
\hline
×                     & ×                    & ×                    & 6D                                & 1,1,1,1,1,1,1,1,1,2          & 1.1           & 1.0           & 2.82                   & 3.10                       \\ 
\hline
                      \checkmark&\checkmark&\checkmark& ×                                 & 2,4,5,8,11,12,13,16,17,24    & 11.2          & 11.5          & \textbf{2.59}          & 29.0                       \\
\hline
\end{tabular}
}
\end{table}

To determine the most effective reward structure for this purpose, we conducted an ablation study on the two primary reward terms responsible for regulating interaction forces: $R_{\text{Force}}$ (penalizing inaccurate fingertip contact forces) and $R_{\text{torque}}$ (penalizing inaccurate joint torques).

Our analysis, summarized in Tab.~\ref{tab:abl_force}, reveals that each reward term shapes the policy's behavior distinctly, and their combination is strictly necessary for achieving robust and adaptive force-modulated grasping:

$R_{\text{Force}}$ Only: Policies produce moderate contact forces and low joint torques. The policy learns to achieve contact but avoids applying high torques, potentially leading to weak or slip-prone grasps.

$R_{\text{torque}}$ Only: Policies achieve moderate joint torques but low contact forces. This suggests the policy learns to ``tense'' the joints without effectively transmitting force through the kinematic chain to the fingertips, resulting in inefficient grasping.

$R_{\text{Force}} + R_{\text{torque}}$: The combination results in significantly stronger and wider ranges for both contact forces and joint torques. This synergistic effect indicates the policy learns to coordinate joint actuation and fingertip contact simultaneously, leading to more forceful, stable, and dynamic grasping.

Furthermore, we evaluated the policy's ability to track a variable force command. As shown in Fig.~\ref{fig:Force_command}, both the exerted contact force and joint torque exhibit an approximately linear relationship with the commanded input, confirming the policy's capacity for fine-grained force control. The strength of this linear correlation directly reflects the efficacy of the reward structure:
\begin{equation}
\text{Correlation Strength: } (R_{\text{Force}} + R_{\text{torque}}) > R_{\text{Force}} > R_{\text{torque}}
\end{equation}
This graded response demonstrates that $R_{\text{Force}}$ is the primary driver for precise force tracking at the end-effector, while $R_{\text{torque}}$ acts as a crucial regularizer that ensures the forces are generated through mechanically efficient and plausible joint-level actuation. Together, they enable the precise, whole-hand force control that is a hallmark of human dexterity.

\subsection{In-Hand Object Rotation}

To quantitatively evaluate the contribution of each sensory component, we conducted a series of real-world ablation studies. Each policy was evaluated over ten trials, with performance measured using three metrics: the number of consecutive successful rotations, average duration per success, and time until failure (see Tab.~\ref{tab:abl_cube}).

The complete observation configuration, comprising force-weighted contact center, contact force, and 6D orientation, has achieved the best performance with an average of 25.1 consecutive successes and a mean duration of 3.36 seconds per success. Replacing the 6D rotation representation with quaternions resulted in a significant performance degradation, reducing the average consecutive successes to 2.9. This sharp decline underscores the importance of a continuous and singularity-free rotation representation for policy learning.

Removing the force-weighted contact center reduced the success count to 12.2, while omitting contact force feedback resulted in 13.2 successes. When both tactile-based components were ablated, performance further decreased to 7.5 consecutive successes. The baseline configuration without any contact sensing performed poorest, averaging only 1.1 successes, which highlights the indispensable role of tactile feedback for in-hand manipulation. Additionally, a policy without explicit orientation feedback achieved 11.2 successes, still underperforming compared to the full observation setup.

These results unequivocally demonstrate that tactile information (i.e., particularly force, weighted contact features and measured contact forces), as well as the 6D object orientation representation are critical for robust and continuous in-hand rotation.

\begin{figure*}[t]
    \centering
    \includegraphics[width=1\linewidth]{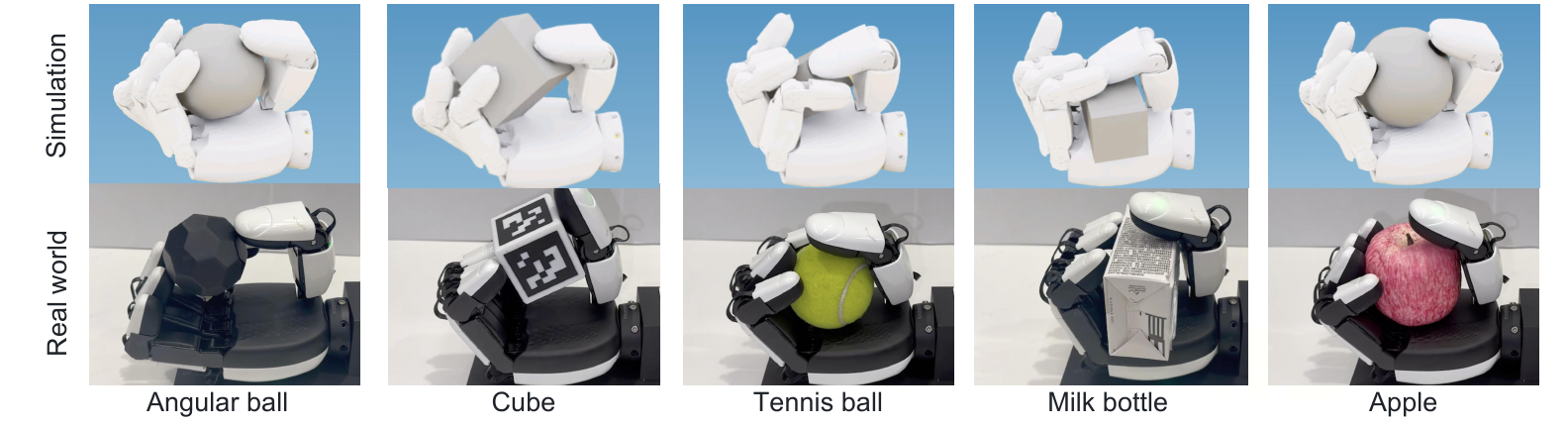}
    \caption{Visualization results of force-adaptive grasping tasks in Real-world and simulation environments}
    \label{fig:grasp}
\end{figure*}

\begin{figure*}[t]
    \centering
    \includegraphics[width=1\linewidth]{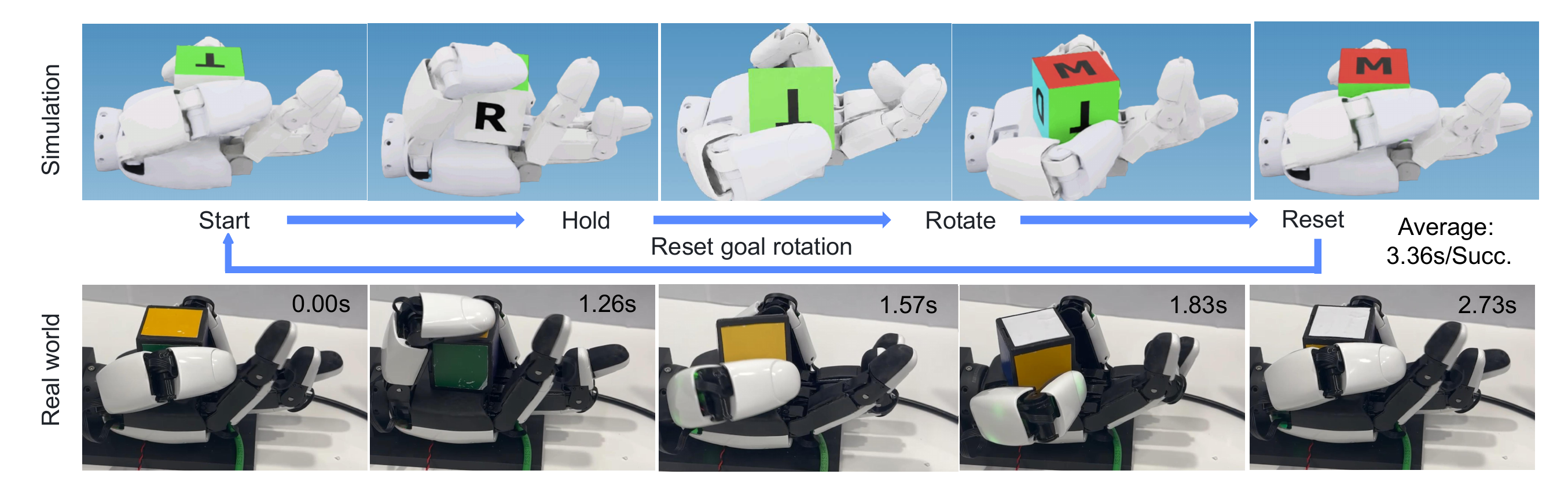}
    \caption{Visualization results of in-hand manipulation tasks in real-world and simulation environments}
    \label{fig:rot_vis}
\end{figure*}

\begin{figure*}[t]
    \centering
    \includegraphics[width=1\linewidth]{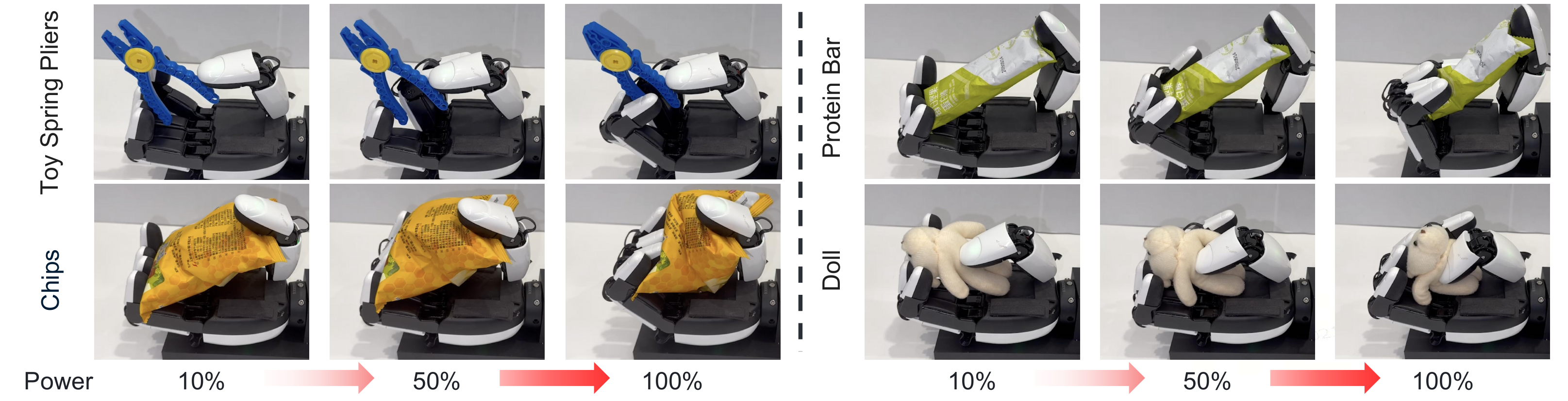}
    \caption{Grasping objects with controllable magnitudes of grasping forces -- from low to high strength.}
    \label{fig:grasp_force_command}
\end{figure*}

\begin{figure}[!thbp]

    \centering
    \includegraphics[width=0.8\linewidth]{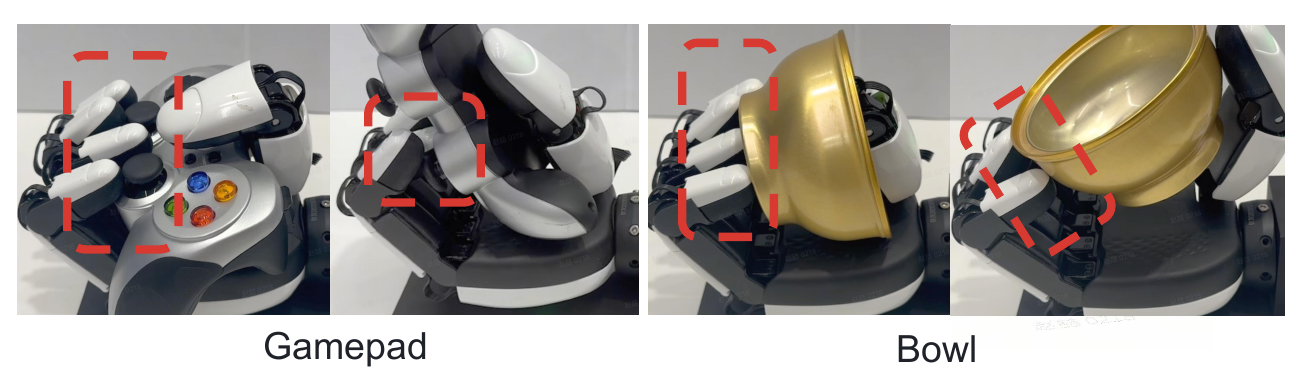}
    \caption{Force-adaptive grasping of irregularly shaped objects that are unseen during training.}
    \label{fig:irregularly}

\end{figure}

\subsection{Simulation and Experimental Results} 

As illustrated in Fig.~\ref{fig:grasp} and Fig.~\ref{fig:rot_vis}, the results from both simulation and real-world experiments are presented for the tasks of force-adaptive grasping and in-hand rotation. A high degree of behavioral consistency is observed between the simulated policy and its real-world execution, demonstrating effective sim-to-real transfer. The policy exhibits remarkably similar motion trajectories and contact patterns across both domains.

Fig.~\ref{fig:grasp_force_command} further demonstrates the response to different force commands, showing that the policy modulates grasp intensity as intended. Moreover, as shown in Fig.~\ref{fig:irregularly}, the policy successfully grasps novel irregular objects not encountered during training, maintaining stable and conforming contact, indicating strong generalization capability to unseen object geometries. This consistency in motion and force response underscores the efficacy of our simulation framework and policy learning approach in bridging the reality gap.

\section{Discussion and Concluding Remarks}
This work successfully demonstrated that robust, force-sensitive dexterous manipulation can be achieved through simulation-trained policies deployed zero-shot on real hardware. Our integrated approach, which combines full-state tactile-torque observations, computationally efficient tactile simulation, and targeted actuator modeling, effectively bridged the primary sim-to-real gaps that have historically plagued dexterous manipulation. The results show that our policies not only perform the tasks of force-adaptive grasping and in-hand rotation but do so with a level of robustness and generalization that is critical for real-world applications.

\textbf{Generalization and Robustness.} A key strength of our method is its ability to generalize. The policy's successful manipulation of unseen, irregularly shaped objects (Fig. ~\ref{fig:irregularly}) indicates that it learned fundamental physical principles of grasping and manipulation, rather than simply memorizing specific object geometries. This emergent robustness is a direct consequence of our extensive domain randomization across contact physics, object properties, and, most importantly, actuator dynamics. By forcing the policy to adapt to a wide spectrum of non-ideal motor behaviors (e.g., randomized torque-speed saturation, backlash), we ensured it would not overfit to a perfect simulated actuator model, thereby achieving remarkable stability on the real hardware.

\textbf{Limitations and Future Work.} Despite its success, our approach has limitations that point toward fruitful future research directions. First, our current tactile simulation, while fast and effective, is a geometric approximation. Integrating a real-time, differentiable soft-body simulator could capture more nuanced contact phenomena like shear forces and multi-directional object slip, potentially unlocking even more delicate manipulation skills. Second, our tasks, though complex, were executed in a structured setting with a fixed hand. The logical next step is to integrate these dexterous hand policies with a mobile manipulator for tasks requiring whole-body coordination and mobility. Finally, our state estimation relied on external vision and onboard IMUs. Learning to perform these tasks directly from raw visual and tactile sensory inputs, without privileged state information, remains a significant but necessary challenge for achieving full autonomy.

\textbf{Broader Impact.} This work serves as a comprehensive handbook for sim-to-real dexterous manipulation. The techniques for current-to-torque calibration and actuator randomization are not limited to the xHand but are applicable to the growing ecosystem of affordable, current-controlled robotic hands. By providing a clear path to bypass the need for expensive joint-level torque sensors and computationally prohibitive tactile simulators, we lower the barrier to entry for research on force-based manipulation.

\textbf{Conclusion.} In conclusion, we have presented a holistic solution to the sim-to-real transfer problem for dexterous, force-controlled manipulation. We showed that by explicitly addressing the gaps in tactile sensing, contact physics, and, most critically, actuator dynamics, it is possible to train simulation policies that execute robustly and effectively on real hardware in a zero-shot manner. 

This study confirms that policies trained with these carefully designed components can execute complex force-adaptive tasks on physical hardware with zero-shot deployment. This framework offers a practical and reproducible \textit{recipe} for robust sim-to-real dexterous manipulation that underpins the wider deployment of dexterous robotic hands to operate autonomously and safely in human-centric environments where fine force adaptation and active regulation of interaction forces are indispensable.

\clearpage

\section{Contributions and Acknowledgements}

\begin{itemize}
    \item Research Ideas and Conceptualization: Haoyu Dong, Zhibin Li, Zhe Zhao.
    \item Training: Haoyu Dong, Zhengmao He, Yang Li, Zhe Zhao.
    \item Evaluation: Haoyu Dong, Zhengmao He, Zhe Zhao.
    \item Hardware Infrastructure Development: Haoyu Dong, Zhe Zhao.
    \item Software Infrastructure Development: Haoyu Dong, Zhengmao He, Yang Li, Xinyu Yi, Zhe Zhao.
    \item Writing: All authors.
    \item Research Direction and Team Lead: Zhibin Li.
\end{itemize}

We express our sincere appreciation to Wenjia Zhu for fostering an emphasis on high‑impact research. The authors thank Yonghui Wu for his strategic insight in the direction of research priorities. The authors also acknowledge the Department Head, Hang Li, for his dedicated support throughout this project.

\clearpage

\bibliographystyle{plainnat}
\bibliography{main}

\clearpage



\end{document}